\documentclass{article}
\usepackage{ijcai25}

\usepackage{times} 
\usepackage[utf8]{inputenc} 
\usepackage[small]{caption} 

\usepackage{soul}
\usepackage{url}
\usepackage[hidelinks]{hyperref}
\usepackage{colortbl}

\usepackage{xcolor}
\colorlet{dark-blue}{blue!50!black}
\hypersetup{
 colorlinks = true,
 citecolor = dark-blue,
 filecolor = dark-blue,
 linkcolor = dark-blue,
 urlcolor = dark-blue
}

\usepackage{amsmath, amssymb, amsthm, bm, mathrsfs}
\usepackage[capitalise]{cleveref}

\usepackage{graphicx}
\usepackage{booktabs}
\usepackage{multirow}
\usepackage{algorithm}
\usepackage{algorithmic}

\usepackage[switch]{lineno}

\title{SOTA: Spike-Navigated Optimal TrAnsport Saliency Region Detection in Composite-bias Videos}

\author{
Wenxuan Liu$^{1,2}$ \and
Yao Deng$^2$ \and
Kang Chen$^1$ \and
Xian Zhong$^{2,}$\footnote{Corresponding author.} \and
Zhaofei Yu$^{3,1}$ \And
Tiejun Huang$^1$ \\
\affiliations
$^1$State Key Laboratory for Multimedia Information Processing, Peking University \\
$^2$Hubei Key Laboratory of Transportation Internet of Things, Wuhan University of Technology \\
$^3$Institute for Artificial Intelligence, Peking University \\
\emails
liuwx66@pku.edu.cn,
361248@whut.edu.cn,
mrchenkang@stu.pku.edu.cn, \\
zhongx@whut.edu.cn, 
\{yuzf12, tjhuang\}@pku.edu.cn
}

\begin{document}

\maketitle

\begin{abstract}
Existing saliency detection methods struggle in real-world scenarios due to motion blur and occlusions. In contrast, spike cameras, with their high temporal resolution, significantly enhance visual saliency maps. However, the composite noise inherent to spike camera imaging introduces discontinuities in saliency detection. Low-quality samples further distort model predictions, leading to saliency bias. To address these challenges, we propose \textbf{S}pike-navigated \textbf{O}ptimal \textbf{T}r\textbf{A}nsport Saliency Region Detection ({SOTA}), a framework that leverages the strengths of spike cameras while mitigating biases in both spatial and temporal dimensions. Our method introduces \textbf{S}pike-based \textbf{M}icro-debias ({SM}) to capture subtle frame-to-frame variations and preserve critical details, even under minimal scene or lighting changes. Additionally, \textbf{S}pike-based \textbf{G}lobal-debias ({SG}) refines predictions by reducing inconsistencies across diverse conditions. Extensive experiments on real and synthetic datasets demonstrate that {SOTA} outperforms existing methods by eliminating composite noise bias. Our code and dataset will be released at \url{https://github.com/lwxfight/sota}.

\end{abstract}

\section{Introduction}

Video saliency detection is essential for isolating objects from backgrounds across consistent frames~\cite{ZhaoLLSZLH24}. With the rapid advancement of digital media, it has become increasingly effective for surveillance applications, particularly in artificial intelligence~\cite{9987642}. However, RGB cameras, constrained by short-exposure shutters, struggle with fast-moving objects and occlusions~\cite{Hu_2022_CVPR}, making continuous and accurate motion capture a significant challenge.


Human visual perception processes motion as continuous and uninterrupted~\cite{2017Cellular}. Inspired by this, neuromorphic cameras abandon the conventional ``frame'' concept and output asynchronous sparse event streams~\cite{10645740}, enabling high-speed object capture independent of shutter speed and frame rate~\cite{zhouguan}. However, they struggle to capture static scenes due to reliance on differential computation~\cite{Zhaorui2024}. In contrast, spike cameras~\cite{7923720} employ a fovea-like sampling method (FSM), simulating the structure and function of the retina. This fusion of motion sensitivity and visual reconstruction allows effective tracking of continuously moving targets.

\begin{figure}[t]
	\centering
	\includegraphics[width = \linewidth]{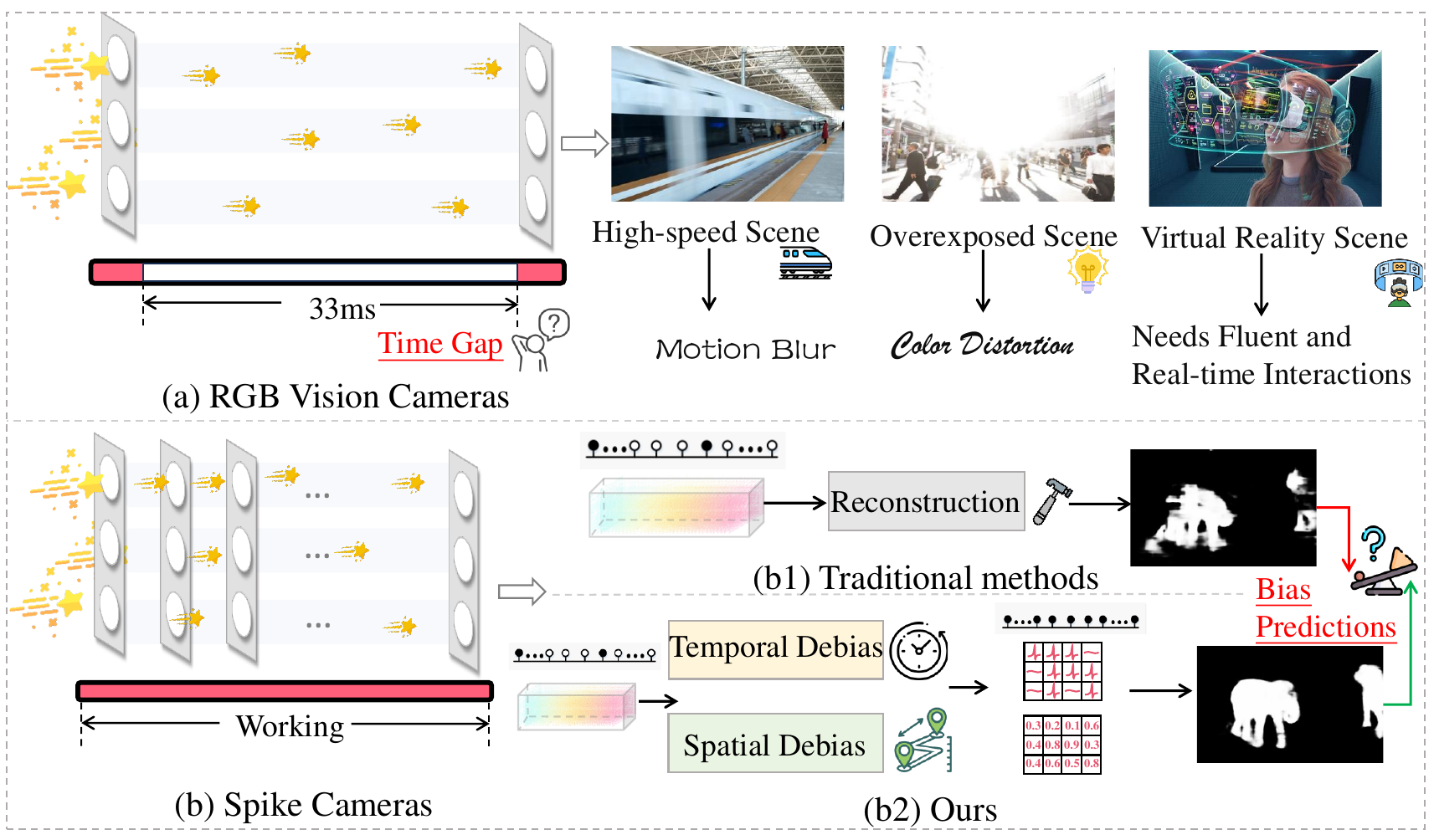}
	\caption{\textbf{Imaging Comparison.} (a) RGB vision cameras, which are affected by time gaps. (b) Spike camera imaging, where (b1) shows the traditional reconstruction method, influenced by environmental noise, leading to inaccurate saliency maps. (b2) presents the optimized reconstruction from a spatiotemporal perspective, resulting in a more accurate saliency map.}
	\label{fig:moti}
\end{figure}

Despite their advantages, both traditional and neuromorphic cameras have limitations in detecting salient regions, as shown in \cref{fig:moti}(a) and (b). Traditional cameras suffer from motion blur, overexposed scenes, and real-time interaction challenges, often losing salient targets. While spike cameras mitigate some of these issues, their reliance on light imaging introduces new challenges. Extended motion over time can cause lighting variations, creating a domain gap in spiking data. A naive temporal transfer approach can introduce noise, amplifying negative effects. This raises the critical question: \textit{How can we restore consistency in target regions while suppressing noise interference?}

\begin{figure}[t]
	\centering
	\includegraphics[width = \linewidth]{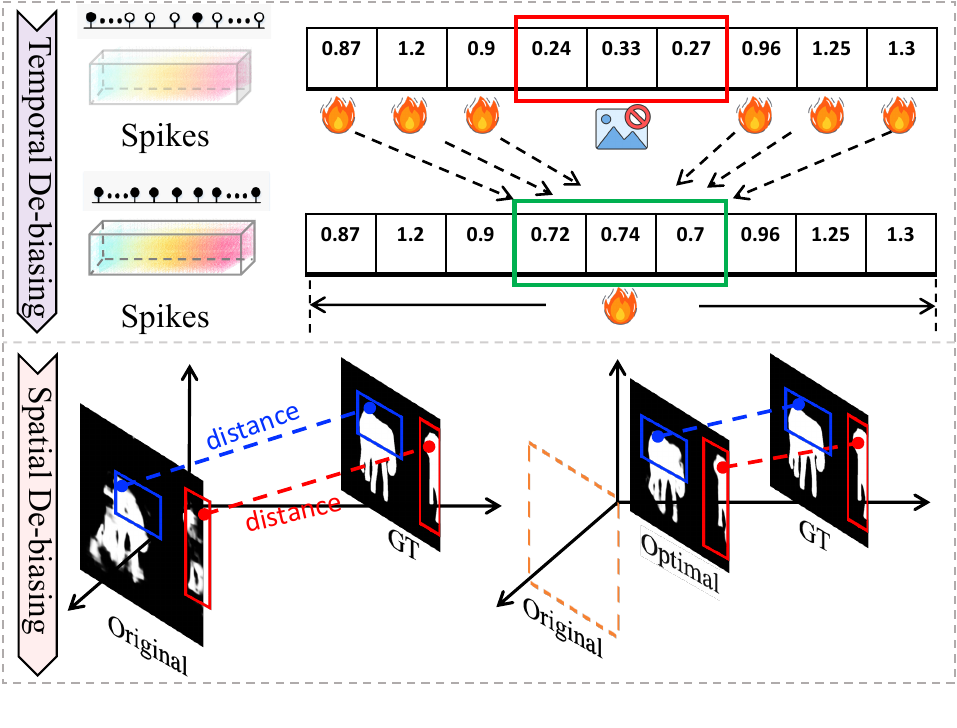}
	\caption{\textbf{Temporal and Spatial Debias.} Temporal debias captures subtle changes by exploring deep feature connections, while spatial debias constructs an OT map to minimize the distance between the spike saliency distribution and the real image distribution.}
	\label{fig:debias}
\end{figure}

To address this, we propose \textbf{S}pike-navigated \textbf{O}ptimal \textbf{T}r\textbf{A}nsport Saliency Region Detection ({SOTA}) in \cref{fig:moti}(b2), leveraging spatiotemporal motion consistency to mitigate composite noise biases. SOTA propagates temporal information between samples to preserve the continuity of high-quality spatial regions, as shown in \cref{fig:debias}. It focuses on two key aspects:

\paragraph{1) Temporal Debias.} To conquer the inconsistency in information among the temporal dimension, we use spiking neural networks (SNNs) to develop a multi-scale strategy for saliency map extraction, which captures deep feature connections across time steps and identifies subtle changes. SNNs generate saliency regions via a threshold mechanism, emitting spikes when the input strength surpasses a predefined threshold, creating a binary sequence. These saliency regions vary across time steps, each representing different confidence levels. We incorporate depthwise separable convolution (DWConv) into the framework to enhance local feature modeling while maintaining efficiency. By capturing deep feature dependencies and minor temporal variations, we improve interactions within the saliency map and mitigate confidence bias.

\paragraph{2) Spatial Debias.} To facilitate the spatial alignment among visual representations, we address spatial saliency correction using an optimal transport (OT) strategy. The core idea is to find the OT map that minimizes the distance between the spike saliency distribution and the real-image distribution. Adversarial learning further refines this mapping to preserve the structural integrity of spike-extracted features. Building on SNN-based temporal bias correction, we extend the spatial subtle changes into temporal-spatial global debias, balancing local details and the global distribution of the spatiotemporal saliency map.

In summary, we introduce {SOTA}, a comprehensive framework for visual saliency detection. {SOTA} extends input features to multi-temporal and multi-scale representations, incorporating a cross-time-step attention mechanism with depthwise convolution for spike-based micro-debias (SM). We formulate saliency map detection as a Kantorovich problem and use adversarial learning to optimize spatial distribution via OT, mitigating spike-based global biases (SG).

Our main contributions are threefold:

\begin{itemize}
	\item \textbf{Novel Modeling.} We bridge traditional imaging and downstream tasks by transforming composite noise in spike streams into domain gap analysis. We propose {SOTA}, which corrects domain bias across both local and global spatiotemporal dimensions, offering new insights for related tasks.

	\item \textbf{Innovative Method.} We use OT to rectify saliency distributions with global contextual information across long-term sequences. Additionally, we introduce {contextual temporal debias}, dynamically inferring positive micro-semantic associations across relevant time steps.

	\item \textbf{Generalization Validation.} We conduct extensive evaluations on real-world and synthetic datasets, thoroughly validating the feasibility and generalizability of {SOTA}. Experimental results in both single-step and multi-step settings demonstrate its effectiveness and confirm the validity of the identified problem.

\end{itemize}





\begin{figure}[t]
	\centering
	\includegraphics[width = \linewidth]{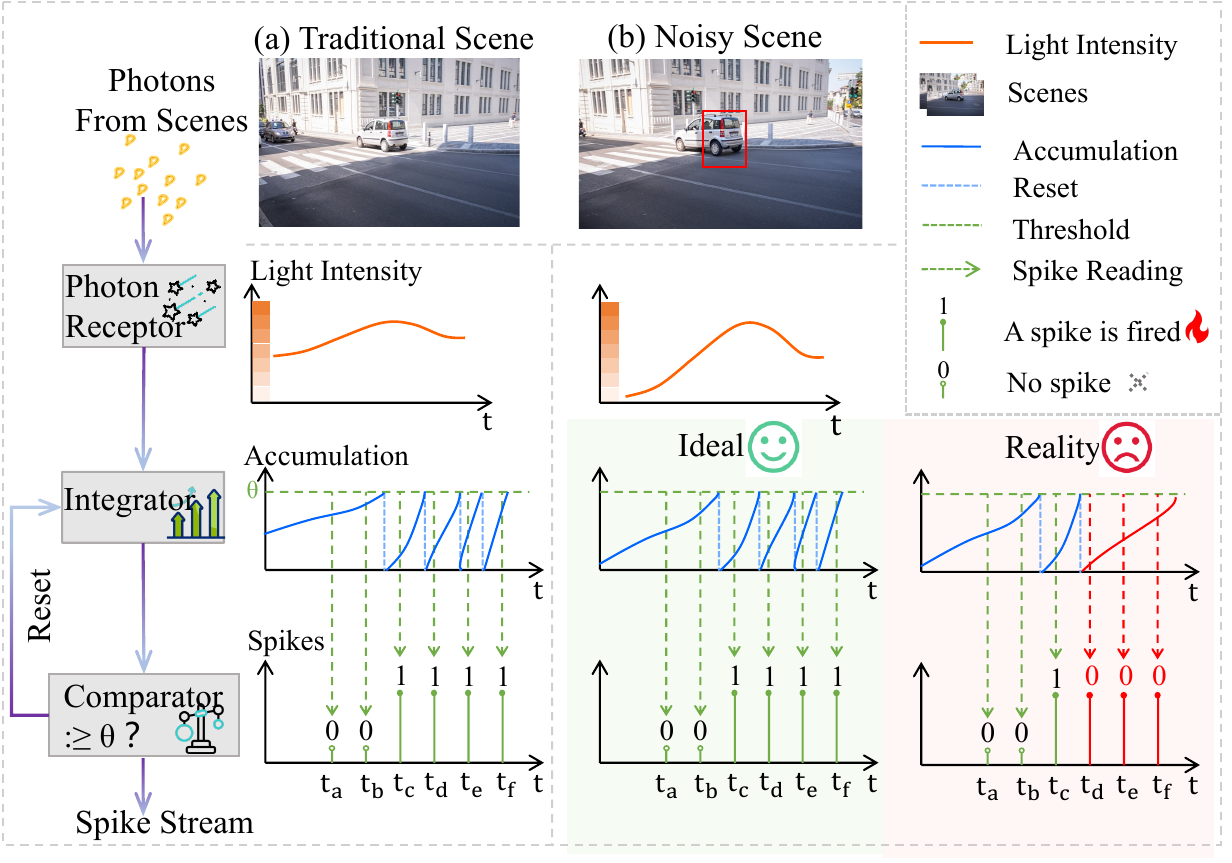}
	\caption{\textbf{Composite Noise Challenges in Spike Cameras.} (a) Traditional imaging forms the foundation of spike cameras. (b) Ideally, continuous spike streams and photons ensure smooth object motion. However, real-world light variations and background interference often result in missing photons, causing biased pixel representations.}
	\label{fig:photons}
\end{figure}

\section{Preliminaries and Motivation}

\paragraph{Preliminaries.}
Unlike event cameras focus on dynamic changes only~\cite{guanMM}, spike cameras simulate the primate retina, with each pixel independently generating spikes in response to variations in light intensity~\cite{huang20231000}. In traditional spike imaging, the CMOS photon receptor converts photons into voltage signals $V$, as shown in \cref{fig:photons}(a). Spikes are generated when the integrated signal reaches a threshold $\Theta$, mapping physical signals to information. The spike generation process is expressed as:
\begin{equation}
	\int_{t_s}^{t_e} I(t) dt \geq \Theta,
\end{equation}
where the voltage signal $V(t)$ at time $t$ is captured from the light intensity $I(t)$ over the temporal interval $\mathcal{T} = [t_s, t_e]$. Spike signals are registered at rates of up to 40 kHz, forming a binary spike stream represented by 0s and 1s.

However, real-world photon accumulation is often imperfect. As shown in \cref{fig:photons}(b), a moving car is divided into two regions, with one half obscured by shadow noise in low-light conditions. Pixels in the shaded region may fail to reach the threshold $\Theta$, preventing spike activation. This leads to reduced {region consistency} of the target object.



\begin{figure}[t]
	\centering
	\includegraphics[width = \linewidth]{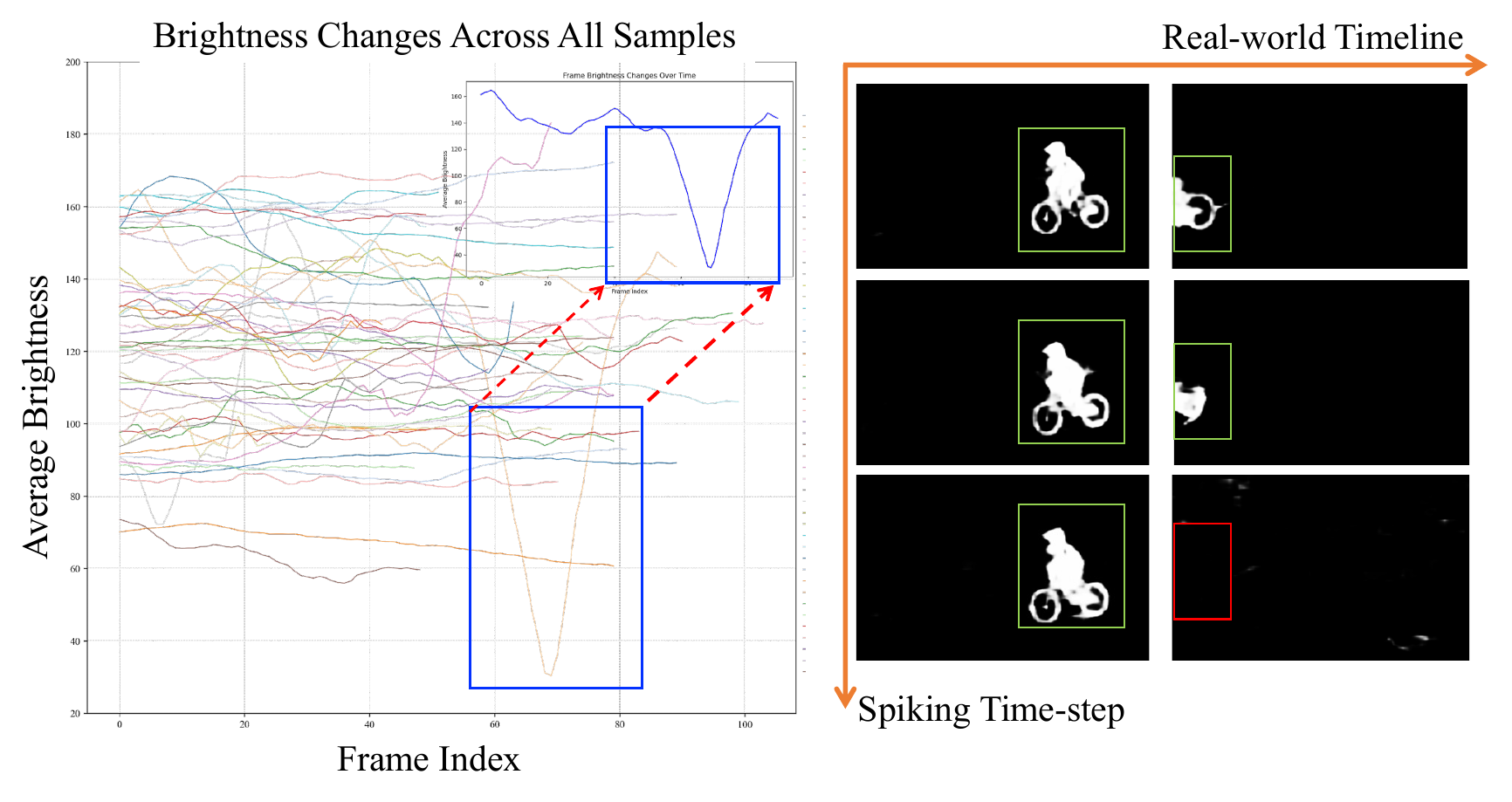}
	\caption{\textbf{Motivation of SOTA.} The left side shows variations in light conditions across samples, with different colored lines representing individual samples. Green boxes indicate correct predictions, while red boxes highlight errors.}
	\label{fig:motivation}
\end{figure}



\paragraph{Motivation.} 
We analyze the temporal variation of light intensity across samples in \textsc{Spike-DAVIS}, as shown in \cref{fig:motivation}. Many samples exhibit significant fluctuations over time. A sample with noticeable ``bumps'' is selected for preliminary experiments, and the results show that RST predictions become inconsistent over time, with performance degrading as brightness decreases. Additionally, multi-step training reveals that information transmitted at different time steps is not always beneficial and may negatively impact predictions.

To mitigate this issue, we propose correcting biases across both spatial and temporal dimensions. First, the model should emphasize details from adjacent time steps in the temporal domain. Depthwise convolution, which processes channel and spatial information separately without increasing computational complexity, is key to introducing local debias.

The optimal transport (OT) problem transforms one distribution into another with minimal cost, either by finding the OT map (Monge problem)~\cite{monge1781memoire} or the OT plan (Kantorovich problem)~\cite{kantorovich2006translocation}. Unlike the Monge problem, the Kantorovich problem considers probabilistic transport, making it a well-defined convex problem with a unique optimal solution.

To strengthen our hypothesis, we reformulate composite bias correction as an OT problem between two probability distributions. The spatial scope of the saliency map is defined by the domains of the source distribution $S$ and target distribution $T$, where $x$ and $y$ are sample points representing pixel-wise saliency scores. Our goal is to refine the initial spike saliency distribution $S$ ($S \in P(X)$) into the target distribution $T$ ($T \in P(Y)$), improving the saliency map’s quality, as shown:
\begin{equation}
	T^*_{\#} S = T,
	\label{eq:tran}
\end{equation}
where the mapping $T^*: Y \rightarrow X$ pushes the initial $S$ to $T$, ensuring the transformed $T^*$ aligns with the target.






\section{Related Works}

\paragraph{Spike Camera-Based Visual Tasks.}
Image reconstruction is fundamental to spike camera-based visual tasks~\cite{chen2022ssml,wgse,tfpTFI,TFSTP}. SpikeGS~\cite{zhang2024spikegs} combines 3D Gaussian Splatting with spike cameras for high-speed, high-quality 3D reconstruction, addressing challenges like motion blur and time-consuming rendering. SpikeNeRF~\cite{Zhu_2024_CVPR} further advances 3D reconstruction by using self-supervision and a spike-specific rendering loss to handle diverse illumination conditions. Spike streams have also been applied to depth estimation tasks, where spatiotemporal embeddings from spike camera streams improve accuracy~\cite{zhang2022spiket}. Additionally, \cite{Zhao_Zhang_Yu_Huang_2024} provides the first theoretical analysis of ultra-high-speed object recognition with spike cameras, proposing a robust representation. Recently, \cite{aaai/ZhuCW024} conducted the first study on visual saliency detection in continuous spike streams, inspiring future spike camera applications in saliency detection.


\paragraph{Sequential Saliency Detection.}
The main challenge in sequential saliency detection is maintaining temporal consistency while capturing smooth target regions in videos. Early methods relied on image-based saliency detection, which failed to capture temporal dynamics~\cite{cvpr/MurrayVOP11}. Later research incorporated temporal information, with notable methods using motion extraction and fusion strategies based on optical flow~\cite{tip/ChenLWQH17}, along with approaches that jointly model spatial and temporal saliency~\cite{guo2024instance}. \cite{zhong2025,actionliu} focus on action regions under multi-view conditions. These advancements significantly improved saliency detection in continuous streams, evolving from static image processing to fully integrated spatiotemporal modeling.

In this paper, we propose the novel {SOTA} framework, which effectively captures local changes and spatial information in temporal streams. Our method optimizes distribution alignment to mitigate domain gaps caused by temporal variations across time intervals.

\section{Proposed Method}

Our goal is to capture and refine smooth salient regions in motion areas reconstructed from the binary spike stream $\mathrm{Spike}(x, y) = \{s(x, y, t)\}$, where $s(x, y, t) = 1$ when a spike is fired. Motivated by the success of Wasserstein GAN~\cite{arjovsky2017wassersteingan}, we frame this process within an adversarial learning framework.

\begin{figure}[t]
	\centering
	\includegraphics[width = \linewidth]{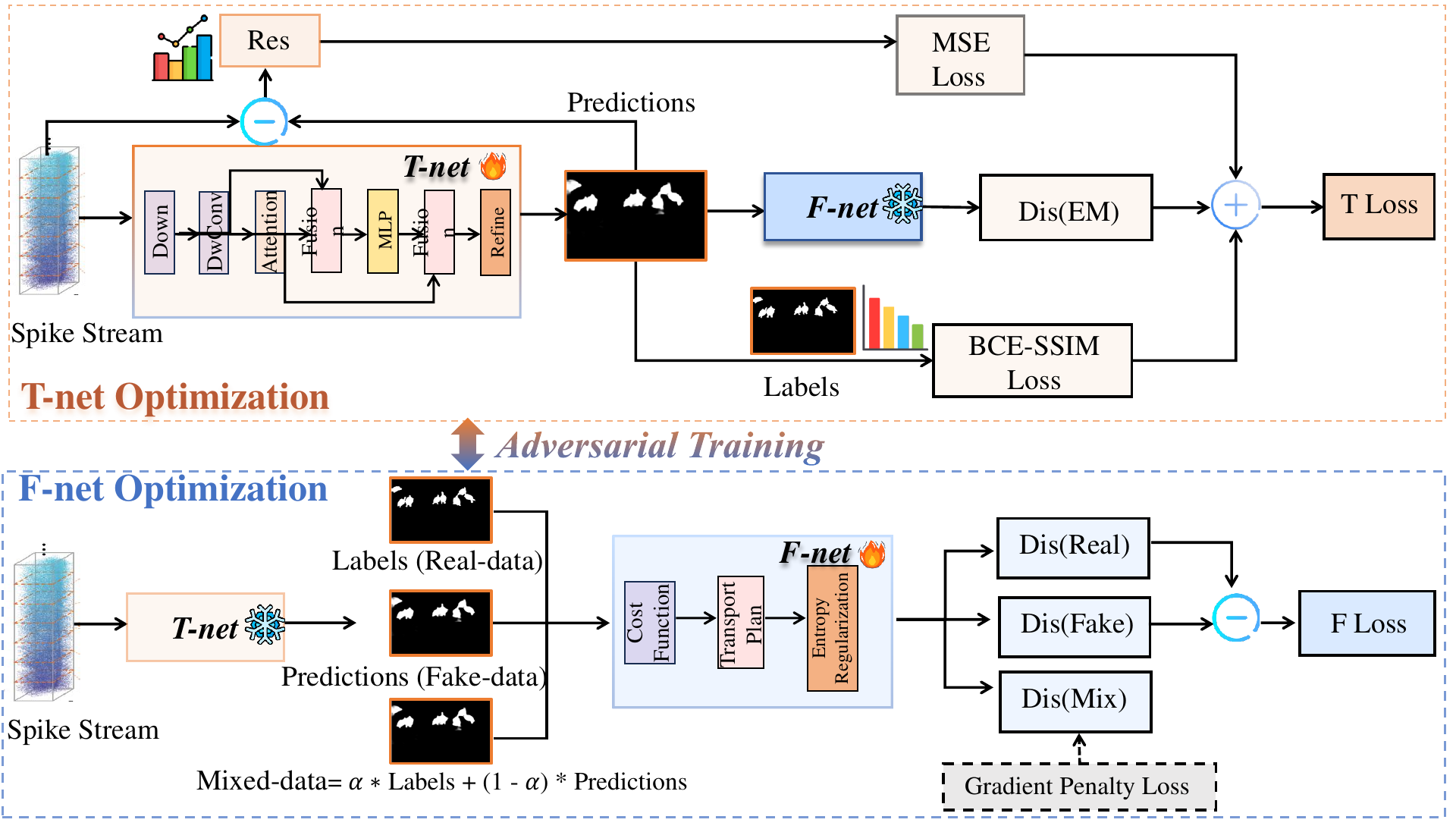}
	\caption{\textbf{Overview of the Proposed SOTA.} The two networks are optimized iteratively: $T$-Net generates temporal saliency maps with micro-detail debias, while $F$-Net refines global spatial debias. Together, they enhance motion associations in the spatiotemporal saliency map and model long-term dependencies.}
	\label{fig:framework}
\end{figure}

{SOTA} takes two inputs: the initial distribution $S$ (sequentially reconstructed from $\mathrm{Spike}(x, y)$) and the target distribution $T$. These are modeled by the Micro-debias network ($T$-Net) and the Global-debias network ($F$-Net), respectively, as shown in \cref{fig:framework}.


\begin{figure}[t]
	\centering
	\includegraphics[width = \linewidth]{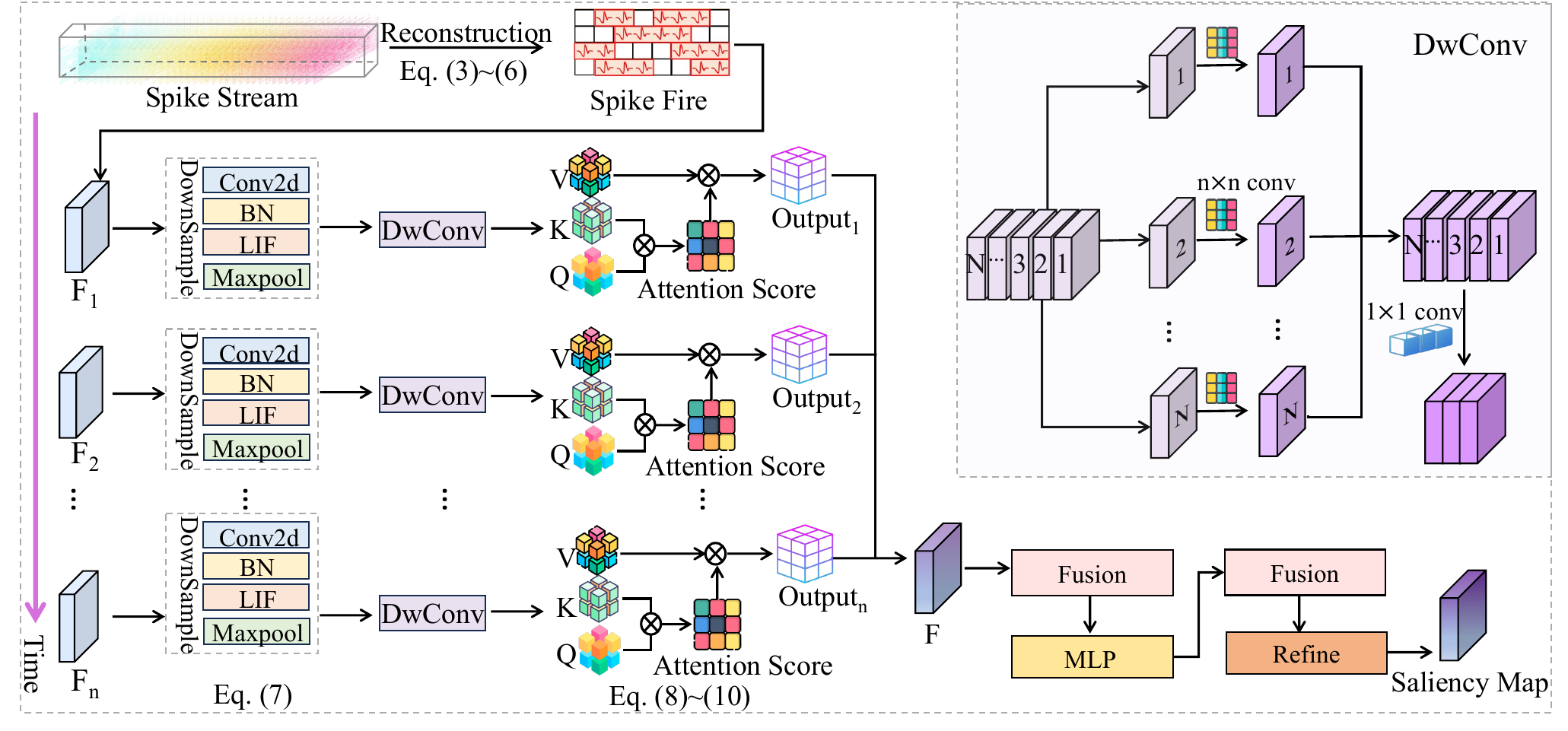}
	\caption{\textbf{Spike-based Micro-debias.} An adjacent-step attention mechanism aggregates temporal features, while DwConv captures fine-grained micro-dynamics.}
	\label{fig:orblock}
\end{figure}

\subsection{Spike-Based Micro-Debias}
\cref{fig:orblock} depicts the spike-based micro-debias (SM) module. Our goal is to enhance positive transfer across time steps and strengthen micro-temporal semantic associations.

\paragraph{Spiking Neuron Firing.}
Unlike continuous signals in traditional artificial neural networks (ANNs), spiking neural networks (SNNs) mimic biological neurons by using discrete spike signals~\cite{xu2023constructing}. The most widely used model is the {leaky integrate-and-fire} (LIF)~\cite{gerstner2014neuronal}, where the input signal influences the neuron's membrane potential $U$. A spike is generated when $U$ exceeds the firing threshold $\theta$, expressed as:
\begin{align}
	\mathrm{Spike}(x, y) = \left\{
	\begin{matrix}
	1, & \text{if~} U > \theta, \\
	0, & \text{otherwise},
	\end{matrix}
	\right.
\label{eq:fire}
\end{align}

The LIF neuron’s event cycle includes a gradual increase in membrane potential due to the input current $C_t$, emission of a spike once the threshold is reached, and a subsequent reset. This process is described as:
\begin{align}
	U &= U[t-1] + C_t, \\
	S &= \tau (U - U_\theta), \\
	V &= U(1 - S) + V_\mathrm{reset} S,
\label{eq:lif}
\end{align}
where $\tau$ is the Heaviside step function, and $U_\theta$ is the firing threshold.

\paragraph{Multi-Scale Spike Feature Modeling.}
To extract multi-scale features effectively, we construct a feature pyramid using successive downsampling:
\begin{align}
	{F}_i = \mathrm{DownBlock}({F}_{i-1}),
\label{eq:pyramid}
\end{align}
where $i = 1, 2, 3, 4$. Each downsampling step reduces spatial resolution while increasing channel depth, yielding a spatial dimension of $\frac{W}{2^i} \times \frac{H}{2^i}$ and a channel size of $C \times 2^i$. The initial feature ${F}$ is extracted from $\mathrm{Spike}(x, y)$ using a CBS block~\cite{aaai/ZhuCW024}, which includes a Conv2d layer, batch normalization, a LIF neuron, and a max-pooling layer.


\paragraph{Learning Multi-Temporal Micro-Interactions.}
Temporal inconsistencies in spike streams~\cite{hsw} can lead to significant performance degradation in SNNs. To mitigate this, we introduce a cross-step operation that reduces bias and enhances interactions across temporal contexts.

We use the final downsampled feature map $F_4$ as input, dividing it into segments $F'_t$ corresponding to time steps $t$. A micro-interactive attention mechanism based on SSA~\cite{aaai/ZhuCW024,SpiKEFormer0023} is then constructed:
\begin{align}
	q &= \mathrm{DwConv}(F'_t, W_Q), \quad \text{for~Query}, \\
	k &= \mathrm{DwConv}(F'_{t+1}, W_K), \quad \text{for~Key}, \\
	v &= \mathrm{DwConv}(F'_{t+1}, W_V), \quad \text{for~Value},
\end{align}
where depthwise convolution $\mathrm{DwConv}$ enables {SOTA} to dynamically adjust spike responses, adapting to subtle sequential variations.

The features are processed using multi-head attention across adjacent time steps~\cite{nips2023Qixu}. The attention result $\mathrm{Att}(F')$ is combined with a projected version of $F'_t$, generating a refined salient feature. The updated feature is further processed through an MLP module, and the outputs are concatenated across time steps for refinement. We adopt the Refine module~\cite{Zhu_2024_CVPR} for this final step.


\subsection{Spike-Based Global Debias}

\paragraph{Kantorovich Modeling.}
The Kantorovich problem (KP) transforms one probability distribution ($\mathbb{S}$, modeled by $T$-Net) into another (the target distribution $\mathbb{T}$) while minimizing the ``transportation cost''. This cost is defined by a function $c(x, y)$, which quantifies the effort required to move mass from one location to another. The problem is formally defined as:
\begin{align}
	\mathrm{Cost}(\mathbb{S}, \mathbb{T}) \overset{\mathrm{def}}{ = } \inf_{\pi \in \Pi(\mathbb{S}, \mathbb{T})} \int_{X \times Y} c(x, y) d\pi(x, y),
	\label{eq:1} 
\end{align}
where the minimum ``transportation cost'' is computed over all transport plans $\pi$, whose marginals correspond to $\mathbb{S}$ and $\mathbb{T}$. The optimal plan $\pi^* \in \Pi(\mathbb{S}, \mathbb{T})$ is the {optimal transport plan}.


\paragraph{Spike-Based Optimal Transport.}
We consider the dual form of KP and adapt it to spike-based optimal transport (OT). The corresponding formulation is:
\begin{align}
	\mathrm{DP\text{-}Cost}(\mathbb{S}, \mathbb{T}) = \sup_{\varphi} \int_Y \varphi^c(y) d\mathbb{S}(y) + \int_X \varphi(x) d\mathbb{T}(x),
	\label{eq:cpst} 
\end{align}
where $\varphi^c(y) = \inf_{x \in X} [ c(x, y) - \varphi(x)]$. The goal is to find the supremum of $\varphi(x)$ and the infimum of $T(y)$. During training, we maximize $\varphi$ and minimize $T$, leading to a max-min adversarial optimization process.

To model this, we use the initial spike distribution $\mathbb{S}$ to approximate $T(y)$ and the target distribution $\mathbb{T}$ using a simple CNN to estimate $\varphi(x)$. Our goal is to optimize Eq.~\eqref{eq:cpst}, where the cost function $c(x, y)$ is a distance metric. The $T$-Net minimizes the discrepancy between $\mathbb{S}$ and $\mathbb{T}$, expressed as:
\begin{equation}
	\int_Y \left[ \tilde{c}(y, T(y)) - \varphi(T(y)) \right] d\mathbb{P}(y).
	\label{eq:RST}
\end{equation}

The $F$-Net minimizes the transport cost between the target distribution and itself while maximizing the transport cost between the spike distribution $\mathbb{S}$ (from $T$-Net) and the target distribution $\mathbb{T}$, as:
\begin{equation}
	\int_X \varphi(x) d\mathbb{Q}(x) + \int_Y \left[- \varphi(T(y)) \right].
	\label{eq:f}
\end{equation}



\subsection{Joint Optimization}

For training, {SOTA} is optimized using two key components. The $F$-Net is updated using the Earth Mover’s (EM) distance and a penalty loss inspired by Wasserstein GAN. The loss function for $T$-Net is defined as:
\begin{equation}
	\mathcal{L}_T = \alpha \mathcal{L}_\mathrm{original} + \mathcal{L}_\mathrm{MSE} - \mathcal{D}_\mathrm{EM},
\end{equation}
where $\mathcal{L}_\mathrm{original}$ follows the RST framework~\cite{aaai/ZhuCW024}, incorporating binary cross-entropy, IoU loss, and SSIM loss. $\mathcal{D}_\mathrm{EM}$ is the EM distance between the $T$-Net output and the target distribution. During inference, the OT processing is not applied.

\section{Experimental Analysis}

\subsection{Datasets and Implementation Details}

\paragraph{Synthetic Dataset.} 
To evaluate the effectiveness of our proposed {SOTA} framework, we construct a bio-inspired video saliency detection dataset, \textsc{Spike-DAVIS}~\cite{cvpr/PerazziPMGGS16}, following prior works on related tasks~\cite{zhang2022spiket}. We use XVFI~\cite{iccv/SimOK21} to interpolate seven images between each frame pair. The dataset includes various actions with challenging attributes such as {fast motion, occlusions, and motion blur}.

\paragraph{Real-World Dataset.} 
\textsc{SVS} dataset, captured using a spike camera with spatial resolution $250 \times 400$ and temporal resolution of 20,000 Hz~\cite{aaai/ZhuCW024}, consists of 130 sequences. Of these, 100 are used for training (24 high-condition, 76 low-condition) and 30 for validation (8 high-condition, 22 low-condition).

\paragraph{Implementation Details.} 
All experiments are conducted on the PyTorch platform with an NVIDIA RTX 4090 24GB GPU. We optimize {SOTA} using the Adam optimizer~\cite{KingmaB14} with an initial learning rate of $2e{-4}$ and a weight decay of $2e{-5}$. The batch size is 2, and input images are resized to $256 \times 256$. For fair comparisons, we apply temporal spike representation and reconstruction on \textsc{SVS} as $M/\Delta t_{x,y}$, following~\cite{aaai/ZhuCW024}, where $M$ is the maximum grayscale value and $\Delta t_{x,y}$ is the spike firing interval at pixel $(x,y)$. We do not fine-tune parameters on \textsc{Spike-DAVIS} and use the same settings as for \textsc{SVS}.

\paragraph{Evaluation Metrics.} 
We evaluate performance using the mean absolute error (MAE), mean F-measure score $mF_\beta$ with $\beta^2 = 0.3$, maximum F-measure $F^m_\beta$, and Structure-measure $S_m$. MAE quantifies overall accuracy, while $mF_\beta$ and $F^m_\beta$ assess precision and recall. The $S_m$ metric integrates object-level and region-level saliency performance. Additionally, we report peak signal-to-noise ratio (PSNR) and structural similarity index (SSIM) to evaluate the impact of image quality.

\begin{table*}[t]
	\centering
 	\small
	\setlength{\tabcolsep}{6pt}
	\begin{tabular}{lc|cccc|cccc}
	\toprule[1.1pt]
	\multirow{2}[2]{*}{Method} & \multirow{2}[2]{*}{Venue} & \multicolumn{4}{c|}{Single Step} & \multicolumn{4}{c}{Multi Step} \\
	\cmidrule(lr){3-6} \cmidrule(lr){7-10}
	& & MAE $\downarrow$ & $F_{\beta}^{\mathrm{max}}$ $\uparrow$ & m$F_\beta$ $\uparrow$ & $S_m$ $\uparrow$ & MAE $\downarrow$ & $F_{\beta}^{\mathrm{max}}$ $\uparrow$ & m$F_\beta$ $\uparrow$ & $S_m$ $\uparrow$ \\
	\midrule
	Spiking Deeplab~\cite{kim2022beyond} & NCE & 0.1026 & 0.5310 & 0.5151 & 0.6599 & 0.0726 & 0.6175 & 0.6051 & 0.7125 \\
	Spiking FCN~\cite{kim2022beyond} & NCE & 0.1210 & 0.4779 & 0.4370 & 0.6070 & 0.0860 & 0.5970 & 0.5799 & 0.6911 \\
	EVSNN~\cite{pami/0012DL0023} & TPAMI & 0.1059 & 0.5221 & 0.4988 & 0.6583 & 0.0945 & 0.6267 & 0.5850 & 0.7023 \\
	Spikformer-OR~\cite{SpiKEFormer0023} & ICLR & 0.1389 & 0.4527 & 0.4408 & 0.6068 & 0.0738 & 0.6526 & 0.6323 & 0.7161 \\
	Spikformer-ADD~\cite{SpiKEFormer0023} & ICLR & 0.1185 & 0.4638 & 0.4415 & 0.6119 & 0.0717 & 0.6890 & 0.6731 & 0.7563 \\
	RST~\cite{aaai/ZhuCW024} & AAAI & {0.0784} & {0.6313} & {0.6171} & {0.6970} & {0.0554} & {0.6981} & {0.6882} & {0.7591} \\ 
	RST~\cite{aaai/ZhuCW024} $\ast$ & AAAI & {0.0776} & {0.6291} & {0.6152} & {0.6968} & {0.0612} & {0.6619} & {0.6515} & {0.7409} \\ 
	\rowcolor{gray!20}
	{SOTA} (Ours) & & \textbf{0.0597} & \textbf{0.6978} & \textbf{0.6880} & \textbf{0.7569} & \textbf{0.0478} & \textbf{0.7402} & \textbf{0.7296} & \textbf{0.7912} \\
	\bottomrule[1.1pt]
	\end{tabular}
	\caption{\textbf{Performance Comparison on \textsc{SVS}.} $\ast$ denotes reproduced results under identical experimental settings. The number of time steps is set to 5.}
	\label{tab_sota}
\end{table*}

\subsection{Comparisons with Existing Methods}
\cref{tab_sota} compares {SOTA} with state-of-the-art SNN-based methods.
Quantitative evaluations are performed under both training settings.
Spiking DeepLab and Spiking FCN~\cite{kim2022beyond}, relying on ANN-SNN conversion, suffer from conversion errors and domain gaps. Other networks struggle to capture dynamic details at each time step, leading to loss of subtle variations in saliency map extraction. Our {SOTA} framework outperforms methods like Spikformer~\cite{SpiKEFormer0023}, EVSNN~\cite{pami/0012DL0023}, and RST~\cite{aaai/ZhuCW024}, achieving an MAE of 0.0478 in multi-step settings.

Spikformer-ADD and Spikformer-OR show improved performance in multi-step conditions, suggesting that they benefit from sequential processing. Additionally, {SOTA} achieves the highest $F_{\beta}^{\mathrm{max}}$ and $mF_{\beta}$ scores, demonstrating its strong ability to balance precision and recall. Notably, {SOTA}'s performance in the single-step setting is slightly lower, highlighting its effectiveness in modeling long-term temporal dependencies.

\begin{table*}[t]
	\centering
	\setlength{\tabcolsep}{7pt}
 	\small
	\begin{tabular}{lccc|cccc|cccc}
	\toprule[1.1pt]
	\multirow{2}[2]{*}{Dataset} & \multirow{2}[2]{*}{\begin{tabular}{c} Base \\ line \end{tabular}} & \multirow{2}[2]{*}{SM} & \multirow{2}[2]{*}{SG} & \multicolumn{4}{c|}{{Single Step}} & \multicolumn{4}{c}{{Multi Step}} \\
	\cmidrule(lr){5-8} \cmidrule(lr){9-12}
	& & & & MAE $\downarrow$ & $F_{\beta}^{\mathrm{max}}$ $\uparrow$ & m$F_\beta$ $\uparrow$ & $S_m$ $\uparrow$ & MAE $\downarrow$ & $F_{\beta}^{\mathrm{max}}$ $\uparrow$ & m$F_\beta$ $\uparrow$ & $S_m$ $\uparrow$ \\
	\midrule
	\multirow{3}{*}{\textsc{Spike-DAVIS}} & $\checkmark$ & & & 0.1119 & 0.3979 & 0.3646 & 0.5969 & 0.0827 & 0.4154 & 0.3974 & 0.6019 \\
	 & \checkmark & & \checkmark & 0.1053 & 0.4191 & 0.3893 & \textbf{0.6091} & \textbf{0.0785} & 0.4179 & 0.3910 & 0.6027 \\
	 & \cellcolor{gray!20}\checkmark & \cellcolor{gray!20}\checkmark & \cellcolor{gray!20}\checkmark & \cellcolor{gray!20}\textbf{0.0861} & \cellcolor{gray!20}\textbf{0.4224} & \cellcolor{gray!20}\textbf{0.4132} & \cellcolor{gray!20}0.6076 & \cellcolor{gray!20}0.0890 & \cellcolor{gray!20}\textbf{0.4281} & \cellcolor{gray!20}\textbf{0.4172} & \cellcolor{gray!20}\textbf{0.6177} \\ \midrule
	\multirow{3}{*}{\textsc{SVS}} & $\checkmark$ & & & 0.0776 & 0.6291 & 0.6152 & 0.6968 & 0.0612 & 0.6619 & 0.6515 & 0.7409 \\
	 & \checkmark & & \checkmark & 0.0649 & 0.6945 & 0.6843 & 0.7539 & 0.0478 & 0.7355 & 0.7240 & 0.7908 \\
	 & \cellcolor{gray!20}\checkmark & \cellcolor{gray!20}\checkmark & \cellcolor{gray!20}\checkmark & \cellcolor{gray!20}\textbf{0.0597} & \cellcolor{gray!20}\textbf{0.6978} & \cellcolor{gray!20}\textbf{0.6880} & \cellcolor{gray!20}\textbf{0.7569} & \cellcolor{gray!20}\textbf{0.0478} & \cellcolor{gray!20}\textbf{0.7402} & \cellcolor{gray!20}\textbf{0.7296} & \cellcolor{gray!20}\textbf{0.7912} \\
	\bottomrule[1.1pt]
	\end{tabular}
	\caption{\textbf{Ablation Study of Components on \textsc{SVS} and \textsc{Spike-DAVIS}.}}
	\label{tab:com1}
\end{table*}

\subsection{Ablation Studies}

\paragraph{Influence of Different Components.} 
We evaluate {SOTA}'s generalization on \textsc{Spike-DAVIS} and \textsc{SVS}, as shown in \cref{tab:com1}. On \textsc{Spike-DAVIS}, the best MAE performance varies between single-step and multi-step approaches. In the single-step setting, {SOTA} achieves the highest $S_m$ score of 0.6091 when the global component is included. This may be due to complex backgrounds in single-step predictions, where feature interactions are insufficient for capturing object-level details. However, introducing $S_m$ leads to a slight MAE degradation by 0.0105.

In multi-step predictions, particularly with strong global dependencies, $S_m$ combined with DwConv may struggle to capture long-range dependencies, leading to incomplete scene representations. In contrast, on \textsc{SVS}, $S_m$ shows more consistent performance, emphasizing its effectiveness in focusing on localized receptive fields.

\begin{table}[t]
	\centering
 	\small
	\setlength{\tabcolsep}{6pt}
	\begin{tabular}{l|cccc}
	\toprule[1.1pt]
	Fusion & MAE $\downarrow$ & $F_{\beta}^{\mathrm{max}}$ $\uparrow$ & m$F_\beta$ $\uparrow$ & $S_m$ $\uparrow$ \\
	\midrule
	OR & 0.0776 & 0.6291 & 0.6152 & 0.6968 \\
	ADD \textit{w/} DwConv & 0.0914 & 0.5978 & 0.5819 & 0.6895 \\
	ADD \textit{w/o} & 0.0758 & 0.6488 & 0.6355 & 0.7171 \\
	SOTA \textit{w/o} DwConv & 0.0649 & 0.6945 & 0.6843 & 0.7539 \\
	\rowcolor{gray!20}
	SOTA & \textbf{0.0597} & \textbf{0.6978} & \textbf{0.6880} & \textbf{0.7569} \\
	\bottomrule[1.1pt]
	\end{tabular}
	\caption{\textbf{Fusion Performance Comparison on \textsc{SVS}.} Results are for single-step training.}
	\label{tab:fusion}
\end{table}

\paragraph{Influence of Micro Debias.}
\cref{tab:fusion} presents the performance of different fusion strategies within SM in the single-step training setting. The OR operation prioritizes overlapping regions between features, achieving an $S_m$ score of 0.6968, but lacks continuity and cumulative effects. In saliency detection, this can lead to inadequate feature fusion, causing loss of critical information or poorly defined salient region boundaries.

ADD fusion integrates multi-modal information stably without excessive noise or complex interactions, but it captures high-level patterns, while DwConv is used for detailed local feature extraction. Combining ADD with DwConv weakens some important patterns, leading to a slight performance decline compared to ADD alone.

\begin{figure}[t]
 	\centering
	\tabcolsep = 1pt
 	\begin{tabular}{cc}
	\includegraphics[width = 0.48\linewidth]{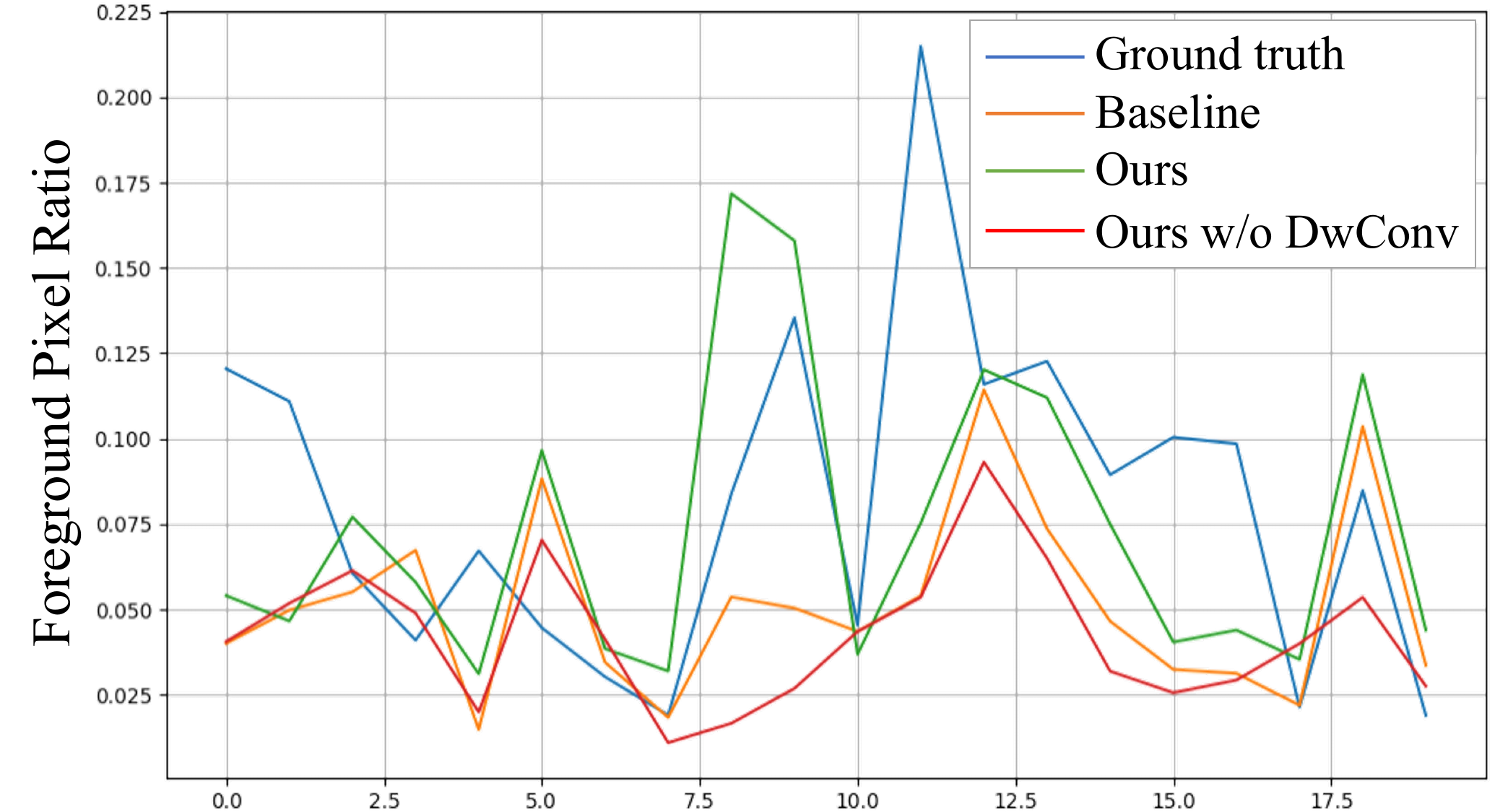} & 
	\includegraphics[width = 0.48\linewidth]{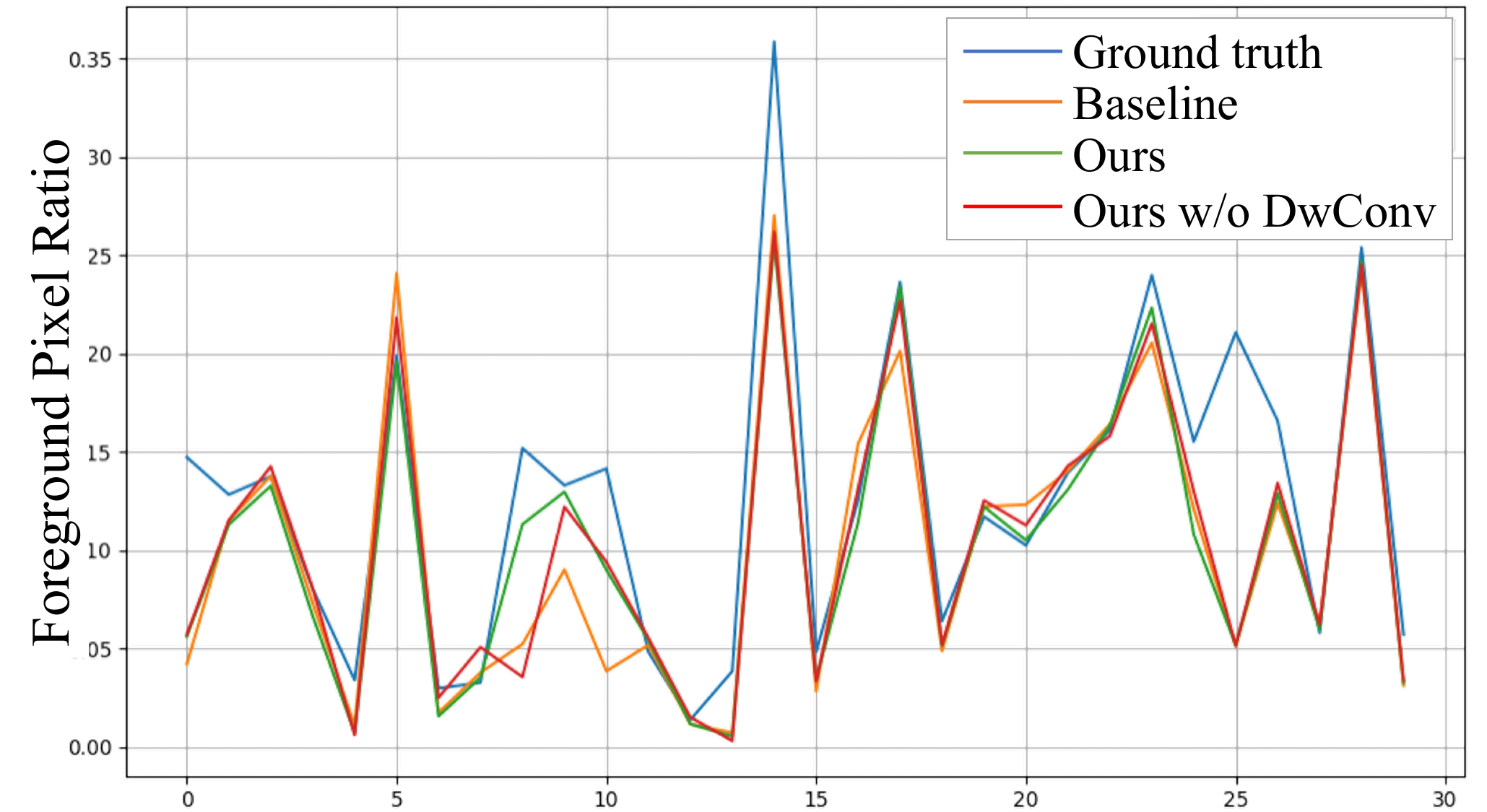} \\
 	\scriptsize{(a) \textsc{Spike-DAVIS}} & \scriptsize{(b) SVS} 
 	\end{tabular}
 	\caption{\textbf{Saliency Pixel Ratio on \textsc{Spike-DAVIS} and \textsc{SVS}.} Colored lines represent different method variants. The $x$-axis denotes the number of categories, and the $y$-axis represents the ratio of action pixels.}
 	\label{fig:pixel}
\end{figure}

\cref{fig:pixel} compares foreground pixel ratios between ground truth (GT), baseline~\cite{aaai/ZhuCW024}, and {SOTA}. In most composite-bias scenarios, {SOTA} outperforms the baseline, reducing background dependency and improving pixel localization. Notably, removing DwConv significantly degrades performance, highlighting the importance of micro-debias. Furthermore, {SOTA} and the baseline show consistent performance in certain classes, suggesting the benefit of exploring adjacent temporal relations for network training optimization.

\begin{table}[t]
 	\centering
 	\small
	\setlength{\tabcolsep}{9pt}
 	\begin{tabular}{l|cccc}
 	\toprule[1.1pt]
 	Distance & MAE $\downarrow$ & $F_{\beta}^{\mathrm{max}}$ $\uparrow$ & m$F_\beta$ $\uparrow$ & $S_m$ $\uparrow$ \\
 	\midrule
 	ED & 0.0933 & 0.5299 & 0.5120 & 0.6408 \\
 	JS & 0.1057 & 0.4988 & 0.4682 & 0.6078 \\
 	KL & 0.0849 & 0.5703 & 0.5494 & 0.6532 \\
	\rowcolor{gray!20}
 	SOTA \textit{w/} EM & \textbf{0.0597} & \textbf{0.6978} & \textbf{0.6880} & \textbf{0.7569} \\
 	\bottomrule[1.1pt]
 	\end{tabular}
 	\caption{\textbf{Ablation Study of Distance Measures on \textsc{SVS}.} Each row presents performance using Euclidean Distance (ED), KL, and JS Divergence, with results for single-step training.}
 	\label{tab:distance}
\end{table}

\paragraph{Influence of Global Debias.} 
{SOTA} optimizes the optimal transport (OT) strategy by measuring the distance between the spike distribution and the target distribution. The choice of distance metric is crucial. We conduct an ablation study on different similarity metrics, with results shown in \cref{tab:distance}.

Kullback-Leibler (KL) and Jensen-Shannon (JS) divergence are effective for evaluating global distribution differences but struggle with local structural information and instability when handling zero probabilities or noise. In contrast, the Earth Mover’s (EM) distance performs better in capturing spatial and sequential structures, which is why we use it for distribution alignment.

\begin{figure*}[t]
	\centering
	\includegraphics[width = \linewidth]{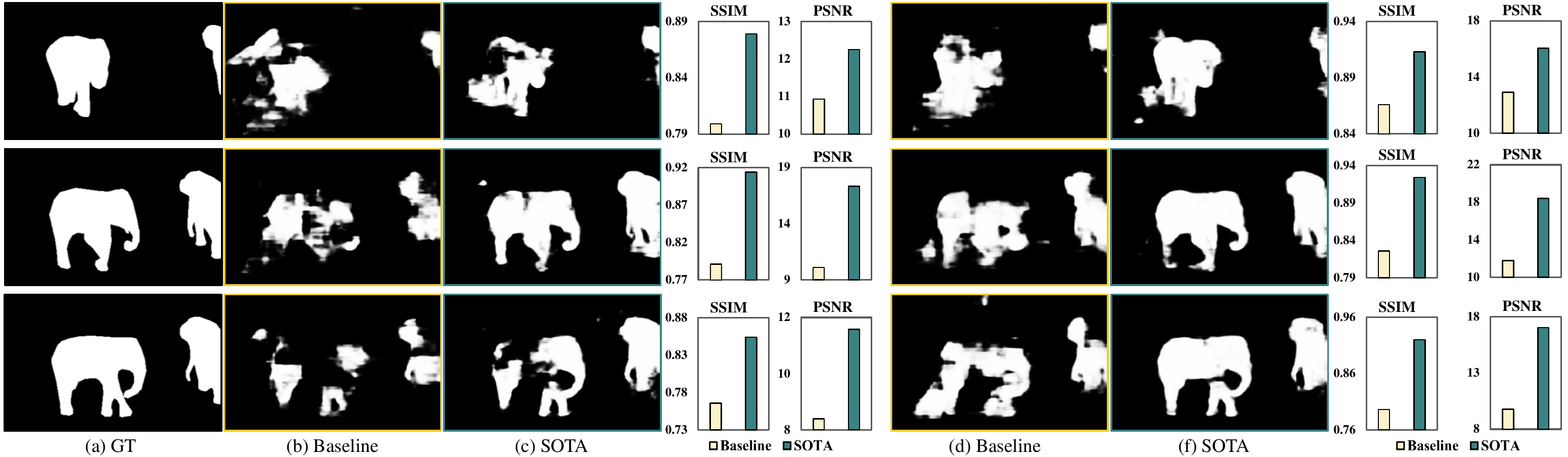}
	\caption{\textbf{Saliency Visualization Results on \textsc{SVS}.} Comparison of final image quality between RST, {SOTA}, and ground truths (GTs). (b) and (c) correspond to single-step training, while (d) and (f) represent multi-step training. Notably, the SSIM and PSNR values for multi-step training are lower than those for single-step training.}
	\label{fig:visual}
\end{figure*}

\subsection{Visualization}

\cref{fig:visual} presents qualitative comparisons across multiple steps on \textsc{SVS}. The bar chart compares image quality metrics, with multi-step predictions outperforming single-step due to enhanced information transfer. {SOTA} shows substantial improvements in prediction accuracy. 

\begin{figure}[t]
	\centering
	\includegraphics[width = \linewidth]{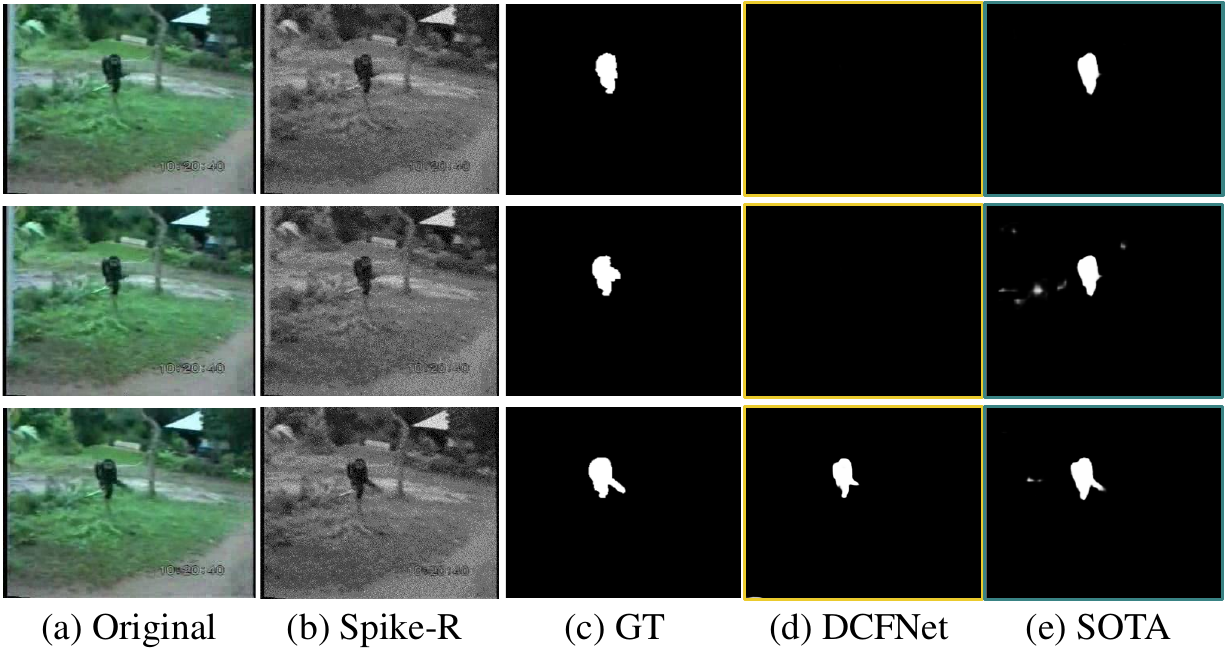}
	\caption{\textbf{Saliency Visualization Results on \textsc{Spike-DAVIS}.} Comparison of final image quality between GTs, CNN-based DCFNet, and {SOTA}. Notably, results are obtained under a multi-step training setting. Spike-R refers to our reconstruction frame.}
	\label{fig:visual2}
\end{figure}

To verify {SOTA}'s effectiveness, we compare it with CNN-based DCFNet~\cite{zhang2021dynamic} on \textsc{Spike-DAVIS} in \cref{fig:visual2}. The DCFNet under our Spike-R setting performs unstably among adjacent frames, while ours clearly identifies saliency regions without losing temporal continuity. However, SOTA introduces slight noise artifacts, which inspired our future work.

\begin{table}[t]
	\centering
 	\small
	\setlength{\tabcolsep}{10pt}
	\begin{tabular}{l|c|c}
	\toprule[1.1pt]
	Method & Single Step & Multi Step \\ 
	\midrule
	RST~\cite{aaai/ZhuCW024} $\ast$ & 1.627 mJ & 14.394 mJ \\
	SOTA \textit{w/o} DwConv & 1.660 mJ & \textbf{13.794} mJ \\
	\rowcolor{gray!20}
	SOTA \textit{w/} DwConv & \textbf{1.589} mJ & 13.872 mJ \\ 
	\bottomrule[1.1pt]
	\end{tabular}
	\caption{\textbf{Efficiency Analysis on \textsc{Spike-DAVIS}.} $\ast$ denotes reproduced results under identical experimental settings.}
	\label{tab:efficient}
\end{table}

\subsection{Energy Efficiency Analysis}
We evaluate the theoretical energy consumption for predicting saliency maps, summarized in \cref{tab:efficient}. Following~\cite{shi2024spikingresformer}, we analyze two variations: {SOTA} with and without DwConv on a GPU to assess energy efficiency. Unlike traditional ANNs, SNNs offer lower power consumption due to their binary nature and reliance on accumulate operations. Our {SOTA} model demonstrates superior energy efficiency compared to traditional methods.

\section{Conclusion}

This paper introduces {SOTA}, a novel method leveraging spike cameras to address the temporal domain gap caused by composite noise in visual saliency detection. The proposed {Spike-based Micro-debias (SM)} module refines temporal transfer across adjacent time steps, ensuring consistent saliency localization and continuous motion capture at the micro level. Meanwhile, the {Spike-based Global-debias (SG)} module enhances the stability of the macro distribution, improving overall visual map accuracy. 
In summary, we integrate adversarial training between these two components to effectively capture local variations and spatial information in sequential data. Additionally, optimal transport is used to align the overall feature distribution, mitigating domain gaps caused by temporal variations between frames. Our method is extensively evaluated on both real-world and synthetic benchmarks, including \textsc{SVS} and \textsc{Spike-DAVIS}, demonstrating superior effectiveness and generalization.


\section*{Acknowledgments}
This work was supported in part by the National Natural Science Foundation of China under Grants 62271361, 62422601, and U24B20140, the Hubei Provincial Key Research and Development Program under Grant 2024BAB039, and the Beijing Nova Program under Grants 20230484362 and 20240484703.


\bibliographystyle{named}
\bibliography{SOTA}

\end{document}